\documentclass[letterpaper, 10 pt, conference]{ieeeconf}  

\IEEEoverridecommandlockouts                              
\usepackage{floatrow}
\newfloatcommand{capbtabbox}{table}[][\FBwidth]
\usepackage{graphicx}
\usepackage{comment}
\usepackage{amsmath,amssymb} 
\usepackage{color}
\usepackage{hyperref}
\usepackage{subfigure}

\usepackage{graphics} 
\usepackage{amsmath} 
\usepackage{amssymb}  
\usepackage{hyperref}
\usepackage{cite}
\usepackage{subfigure}
\usepackage{graphicx}
\usepackage{setspace}
\usepackage{multirow,makecell}
\usepackage{xcolor}
\title{\LARGE \bf
SRH-Net: Stacked Recurrent Hourglass Network for Stereo Matching
}

\author{Hongzhi Du$^{*1}$, Yanyan Li$^{*2}$, Yanbiao Sun$^{\dagger1}$, Jigui Zhu$^{1}$ and Federico Tombari$^{2,3}$ 
\thanks{$*$ authors are with equal contributions; $\dagger$ Corresponding author}
\thanks{$^{1}$State Key Laboratory of Precision Measuring Technology and Instruments, Tianjin University, 300072, Tianjin, China
        {\tt\small (duhz, yanbiao.sun, jiaguizhu)@tju.edu.cn}}%
\thanks{$^{2}$Dept. Computer Science, Technical University of Munich, Munich, Germany
        {\tt\small (yanyan.li, federico.tombari)@tum.de}}%
\thanks{$^3$ Google Inc.}        
}

\begin{document}

\maketitle
\thispagestyle{empty}
\pagestyle{empty}

\begin{abstract}
The cost aggregation strategy shows a crucial role in learning-based stereo matching tasks, where 3D convolutional filters obtain state of the art but require intensive computation resources, while 2D operations need less GPU memory but are sensitive to domain shift. 
In this paper, we decouple the 4D cubic cost volume used by 3D convolutional filters into sequential cost maps along the direction of disparity instead of dealing with it at once by exploiting a recurrent cost aggregation strategy.
Furthermore, a novel recurrent module, Stacked Recurrent Hourglass (SRH), is proposed to process each cost map. Our hourglass network is constructed based on Gated Recurrent Units (GRUs) and down/upsampling layers, which provides GRUs larger receptive fields. Then two hourglass networks are stacked together, while multi-scale information is processed by skip connections to enhance the performance of the pipeline in textureless areas. 
The proposed architecture is implemented in an end-to-end pipeline and evaluated on public datasets, which reduces GPU memory consumption by up to 56.1\% compared with PSMNet using stacked hourglass 3D CNNs without the degradation of accuracy.
Then, we further demonstrate the scalability of the proposed method on several high-resolution pairs, while previously learned approaches often fail due to the memory constraint. The code is released at \url{https://github.com/hongzhidu/SRHNet}.

\end{abstract}

\section{INTRODUCTION}
Stereo matching aims to recover the dense reconstruction of unknown scenes by computing the disparity from rectified stereo images, helping robots intelligently interact with environments. To improve the accuracy of disparity estimation, computationally intensive methods are performed by geometric~\cite{988771} and learning-based approaches~\cite{Zhang2019GANet}. Those methods, however, are limited to only run on devices with powerful computing capabilities. Hence, improving the accuracy of stereo matching algorithms within limited computation resources is still an open topic.

The traditional processing of stereo matching involves three components: cost computation, cost aggregation and disparity computation~\cite{988771}. Instead of using hand-craft functions in this pipeline, learning cost computation~\cite{Zbontar2015} or regularization~\cite{Li2008,Scharstein2007} approaches are merged into traditional stereo algorithms, whereas they are still insufficient for correct matching in textureless, occluded and repetitive patterned scenes. Different from those mixed approaches, end-to-end architectures aim to directly estimate disparity from stereo images to improve the robustness in those challenging regions. 

According to the usage of 3D convolutions in stereo matching, those networks are classified into two groups: encoder-decoder networks based on 2D convolutional operations (\textbf{2DCoder})~\cite{kang2019context,wang2020fadnet} and explicit cost aggregation networks with 3D convolutional filters (\textbf{3DConv})~\cite{kendall2017end,chang2018pyramid}. 
The earliest 2DCoder method can date back to the DispNet~\cite{mayer2016large}, in which a 1D correlation layer is used to generate a 3D cost volume. By concatenating with a feature map, this volume can be conducive to the disparity regression in the process of decoding. Based on the DispNet, residual structures are used in CRL~\cite{pang2017cascade} and FADNet~\cite{wang2020fadnet} to implement a deep network and refine the disparity across multiple scales. These methods represent matching costs in unary values to finish aggregating a 3D volume by 2D convolutions. To avoid the collapse of feature dimension during cost aggregation, 4D cost volumes are constructed in 3DConv methods. GC-Net~\cite{kendall2017end} firstly proposes the basic framework for such approaches that include feature extraction, cost aggregation/regularization, and disparity regression. To improve the accuracy, more intensive aggregation modules are proposed, such as the stacked hourglass 3DCNN in PSMNet~\cite{chang2018pyramid}, guided aggregation layers in GA-Net~\cite{Zhang2019GANet}, and the neural architecture search framework in LEAStereo\cite{cheng2020hierarchical}. However, these 3D operations for 4D cost volume aggregation bring huge computation and require a large amount of GPU memory for employment.
\begin{figure}[t]
  \centering
  \subfigure[Input image]{
   \includegraphics[width=0.46\columnwidth]{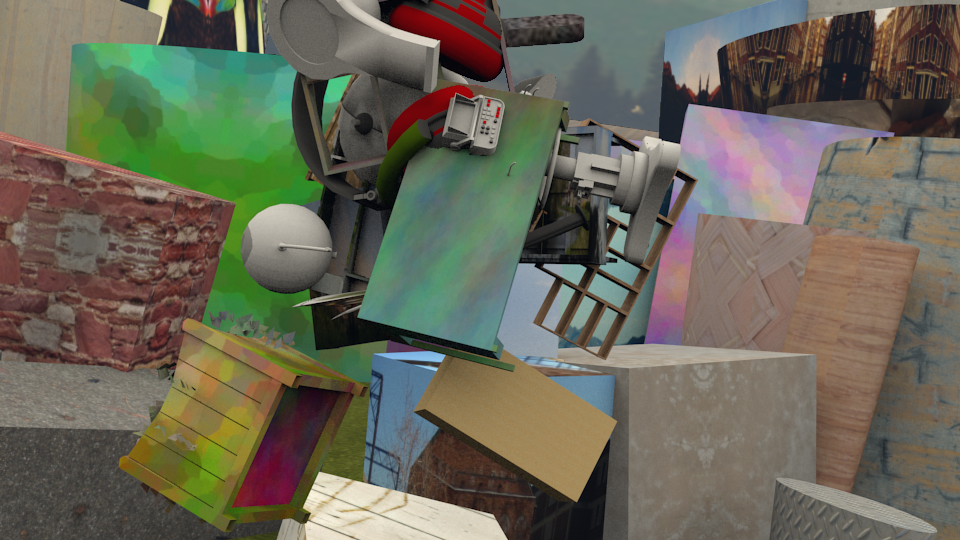}
   } 
   \subfigure[Stacked GRU~\cite{yao2019recurrent}]{
   \includegraphics[width=0.46\columnwidth]{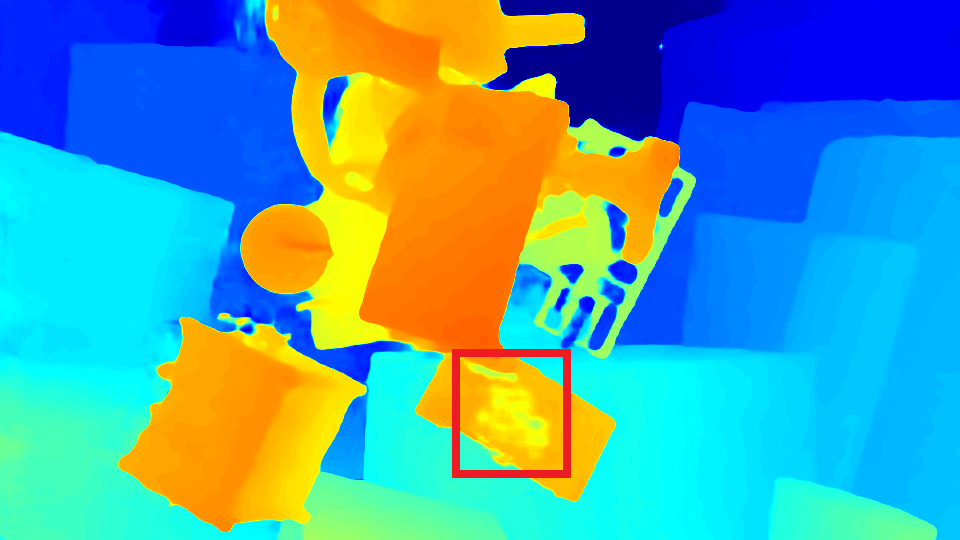}
   }\vspace{-2mm}
  \subfigure[SRH-Net]{
   \includegraphics[width=0.46\columnwidth]{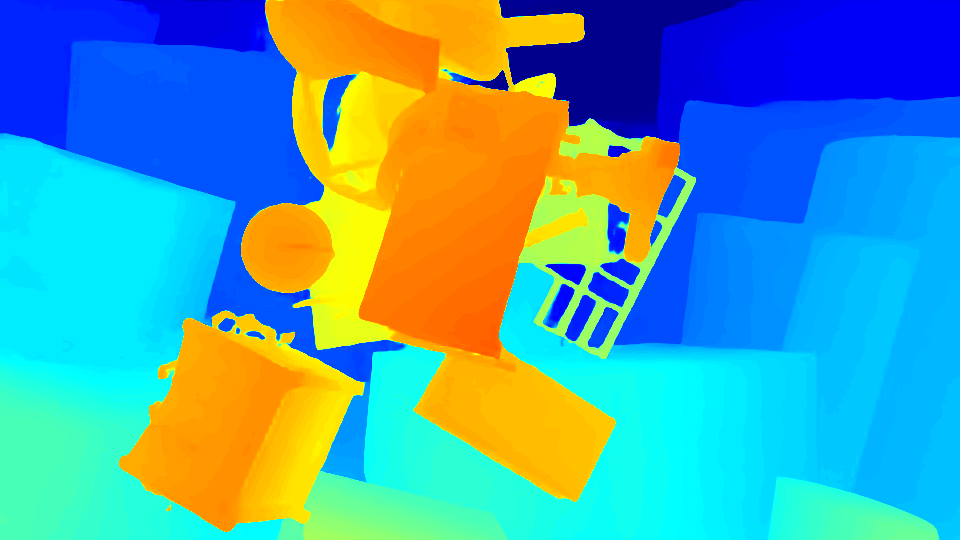} 
   }
   \subfigure[Ground truth]{
   \includegraphics[width=0.46\columnwidth]{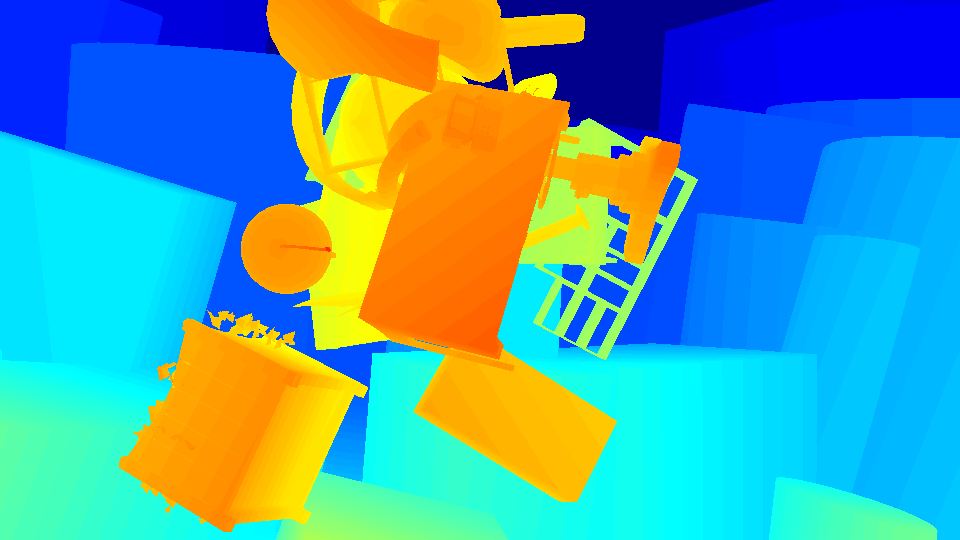}
   }
   \caption{Performance of our SRH module embedding into cost aggregation. Our SRH-Net performs more robust disparity estimation in textureless regions than the stacked GRU method used in RMVSNet~\cite{yao2019recurrent}.}
      \label{fig:example}
\end{figure}

To alleviate the consumption of computation resources, correlation layers are used in the pipeline of 3DConv methods to reduce the channels of the cost volume~\cite{gwc2019,AANet}. However, a low-dimensional cost representation, which breaks the integrity of information, cannot achieve the performance as the concatenation volume. 

Inspired by the performance of stacked RNN layers used in multi-view depth prediction tasks~\cite{yao2019recurrent}, the 4D cost volume in this paper is also processed sequentially. But different from the traditional stacked RNNs~\cite{yao2019recurrent} that aggregates cost maps under a single scale, the proposed SRH module generates a cost pyramid composed of three-scale maps, and then each hourglass is repeated by bottom-up and top-down processing to improve the robustness of the network, especially in occluded and low-textured areas.
 
As shown in Figure~\ref{fig:example}, our SRH module performs better than the stacked GRU method with the same feature extraction block. Thanks to our approach, we obtain comparable accuracy on public datasets, while reporting a GPU memory consumption which is just \textbf{32.1\%} of GA-Net~\cite{Zhang2019GANet} and \textbf{43.9\%} of PSMNet~\cite{chang2018pyramid}. The main contributions of this paper are summarized as follows:
\begin{itemize}
\item An efficient recurrent cost aggregation strategy is used to take place of 3D convolutional filers in stereo matching.

\item A novel cost aggregation module leveraging GRUs with convolutional layers is proposed to capture multi-scale information at each disparity level.

\item Our end-to-end pipeline obtains competitive results in high-resolution reconstruction with using limited memory.
\end{itemize}

\begin{figure*}[t]
  \centering
   \includegraphics[width=\linewidth]{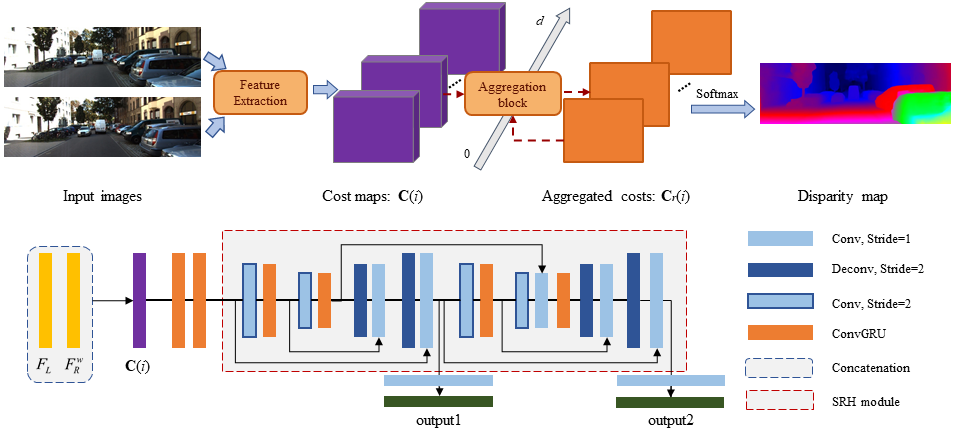}
   \caption{Overview of the proposed SRH-Net (above) and architecture of the cost aggregation block(below).}
   \label{fig:Architecture}
\end{figure*}

\section{RELATED WORK}
Traditional methods relying on matching algorithms extract corresponding points or patches between two images, and then correct wrong matches by using cost aggregation approaches~\cite{Hosni2013Fast,Hirschmuller2008Stereo}. However, they are sensitive to ill-posed areas because of hand-crafted functions. To improve accuracy and robustness in textureless and occluded regions, end-to-end learning approaches are proposed including 2DCoder and 3DConv. Furthermore, we also illustrate the recurrent cost aggregation adopted in this paper.


\textbf{Encoder-decoder}
DispNet~\cite{mayer2016large} proposes an encoder-decoder architecture for disparity estimation, which captures large disparity by downsampling operations first, then upsampling the feature maps to $384\times192$ in the decoder. In DispNetC, expanded from DispNet, a 3D cost map is generated by a correlation operation to cover a maximum disparity of 40 pixels. Then those costs and features are regressed to a disparity map in the process of decoding. Inspired by the classical encoder-decoder architecture, CRL~\cite{pang2017cascade} provides a two-stage pipeline in a cascade manner, while FADNet~\cite{wang2020fadnet} constructs multi-scale disparity predictions by building two parallel encoder-decoder modules. Both of them exploit residual structures to make their models deeper and make initial disparities more smooth. To improve the feature extraction performance of DispNet, CPNet~\cite{kang2019context} encodes multi-scale features by using dilated convolution operations with different dilation rates, and also proposes a gradient regularizer in the training process.

\textbf{3D convolutional networks}
Different from pipelines introduced in the last section, methods in the 3DConv class have an explicit cost aggregation step. In GC-Net~\cite{kendall2017end}, feature maps of left and right images are extracted by a weight-sharing siamese block and used to construct the 4D cost volume. After aggregating this cost volume by 3D convolutions, a soft argmin function is applied to regress the final disparity result. On this basis, PSMNet~\cite{chang2018pyramid} employs a sparse pyramid pooling module during feature extraction to build the cost volume that is captured from global context information. In the regularization stage, the stacked hourglass 3D CNN is used to aggregate the volume with intermediate supervision. Although PSMNet improves accuracy by using these strategies, the cubic computational burden is brought to the network at the same time. GA-Net~\cite{Zhang2019GANet} proposes two novel layers, semi-global guided aggregation (SGA) layer and local guided aggregation (LGA) layer, to achieve a better result in textureless regions and thin structures. In its cost aggregation stage, apart from guided aggregation layers, sub-volumes generated in the cost aggregation stage need to be connected by concatenation operations, this bringing additional GPU memory burden than element-wise additions in GC-Net and PSMNet.

\textbf{Recurrent cost aggregation}
GRU~\cite{elman1990finding,cho2014learning} is a type of recurrent neural networks (RNNs)~\cite{wang2011action,donahue2015long} proposed for the purpose of exploiting sequential information. In multi-view depth prediction tasks that try to predict the depth map from different views, RMVSNet~\cite{yao2019recurrent} replaces 3D CNNs used in MVSNet~\cite{yao2018mvsnet} with stacked convolutional GRUs. Compared with strategies~\cite{gwc2019,AANet} that reduce channels of the cost volume by using the correlation operation, the RNNs-based methods remain the complication of the cost volume by splitting it into cost maps $C_j, j\in (0, d)$. At the $i^{th}$ iteration of aggregation step, the GRU unit is fed by $C(i)$ and $C_r(i-1)$, where $C(i)$ presents the unaggregated cost while $C_r(i-1)$ shows the aggregated cost generated from the $(i-1)^{th}$ iteration. 


As a direct strategy for enhancing the property of the recurrent network, a 3-layer stacked GRU module is exploited in RMVSNet~\cite{yao2019recurrent} to regularize cost maps. However, the traditional stacked RNNs are still not suitable for dense stereo tasks since they cannot capture multi-scale information. 
To enhance the performance of recurrent networks in stereo matching tasks, our module stacks hourglass networks instead of GRU layers directly, where each cost map is aggregated at different scales benefitting from added convolutional layers in each hourglass network.

\section{METHOD}

The cost aggregation strategy is useful for end-to-end networks to refine the matching results, although 2DCoder pipelines cannot aggregate matching cost sufficiently while 3D convolutional modules consume too much computation. In this section, a novel pipeline, SRH-Net, is introduced to solve these issues, by building cost pyramids for each cost map along the disparity direction. As shown in Fig.~\ref{fig:Architecture}, the whole network consists of three main blocks: cost map construction, cost aggregation and disparity regression.

\subsection{Cost map construction}
The deep features of stereo images are extracted separately by a 2D siamese block where weights are shared in both branches. Similar to PSMNet~\cite{chang2018pyramid}, we also adopt the spatial pyramid pooling (SPP) module~\cite{he2015spatial} in feature extraction to incorporate hierarchical context relationship. After obtaining the feature maps, a sequence of cost maps is constructed by concatenating the left feature map $F_L$ with the warped right feature map $F_R^w$ at the specified disparity level. Since the input stereo images have been rectified, correspondences are localized on the same y-axis line in the left and right images. Given a disparity searching level $i$, the right feature map $F_R(u,v)$ can be shifted to $F_R^w(i)$ following 
\begin{equation}
    F_R^w(i) = {F}_{R}(u-i,v)
\end{equation}
where $(u, v)$ is a 2D position in the right feature map, which is moved to a new position $(u-i,v)$ in the warped image. 
Then, the cost map $C(i)$ is constructed via the following step, 
\begin{equation}
    C(i)=\left[ F_L, F_R^w(i)\right]
\end{equation}
where $[\cdot,\cdot]$ denotes the concatenation operation. In our experiments, we have $64$ cost maps along the disparity direction. Those cost maps, from $C(0)$ to $C(d)$, are fed to our cost aggregation block sequentially as shown in Figure~\ref{fig:Architecture}.

\subsection{Recurrent cost aggregation}\label{section:recurrent aggregation}
Different from the 3DConv approaches that aggregate 3D cost volume directly, the recurrent cost aggregation achieves sequential processing with fewer parameters and GPU memory. As shown in figure~\ref{fig:Architecture}, we apply a cascade of two GRU layers at the beginning of this block as a pre-procession stage, which merges the feature in cost maps. Then, there are two recurrent hourglass modules following the two GRUs, which capture more context information for the case of low-textured areas and repetitive patterns (more detailed information are described in ~\ref{section:SRHmodule}). 

The aggregation process generates two 1-channel cost maps (`output1' and `output2') at the current disparity level via convolutional layers. In the training phase, 'output1' is used for the intermediate supervision~\cite{wei2016} while `output2' is applied to predict the ultimate disparity map for evaluation. Finally, the proposed block is fed by each cost map iteratively along the direction of disparity arrange to deal with all matching cost maps as shown in Figure~\ref{fig:Architecture}.

\subsection{Stacked recurrent hourglass module}\label{section:SRHmodule}


Following the main idea of sequential processing, a SRH module is proposed to address issues in the stacked, improving the performance of 2D-based recurrent networks for stereo matching by means of an effective SRH module during cost aggregation. Differently to the standard stacked hourglass~\cite{Newell2016Stacked}, the SRH module is composed of 2D CNN and GRU layers, where cost maps are downsampled to different scales and merged in an end-to-end pipeline for our recurrent cost aggregation stage.
 

In each sub-module, GRU layers are used after the convolutional layers with 2 strides, then after reaching the lowest resolution the module starts the decoding process by means of two deconvolutional operations. In this way, sub-modules can capture more context information for cost aggregation and retrain the resolution of $C(i)$ in outputs.

Then, the SRH module is composed by stacking these two recurrent hourglasses end-to-end, feeding the output of the first hourglass as input for the next one. The second hourglass will further increase the receptive field and makes our network deeper.
To consolidate information across scales, each cost map in the cost pyramid is processed again to be further aggregated based on skip connections.

Compared with the stacked GRU layers that only capture cost information at a single scale, our design builds a cost pyramid at each disparity level,
where cost maps are processed into different scales. Here, fine-grained information is contained in the bottom part of the cost map since the convolutional kernels capture smaller areas. In the meantime, structural and contextual information is learned in the top part of the cost map, where convolutional kernels provide larger receptive fields. Furthermore, skip connections in our SRH module provide chances for information streaming across multi-scale cost maps, which also enhance the robustness of our network in dealing with textureless areas, blur and occlusions compared with traditional stacked GRU structures.       

\begin{figure*}[t]
\centering
\subfigure[Input]{
\begin{minipage}[b]{0.24\textwidth}
\centering
\includegraphics[width=\textwidth]{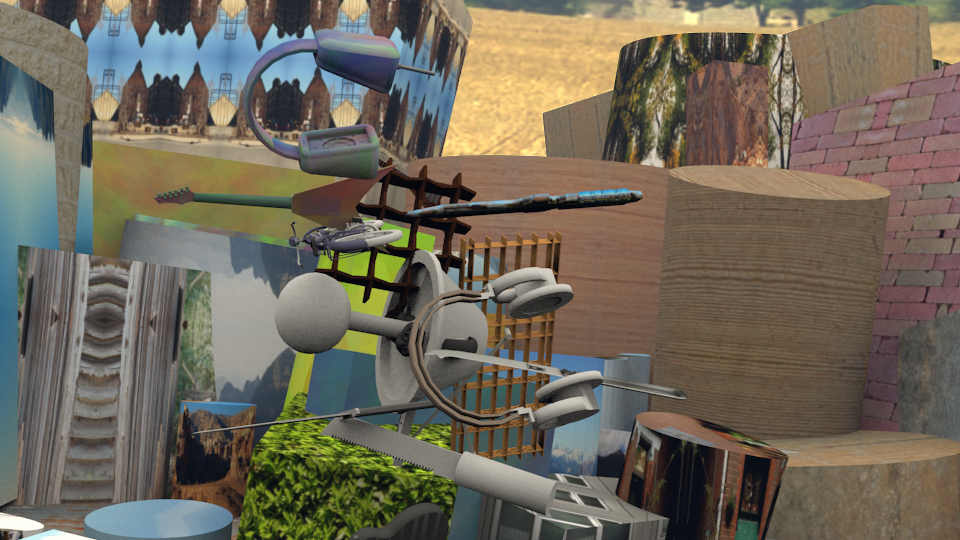}
\includegraphics[width=\textwidth]{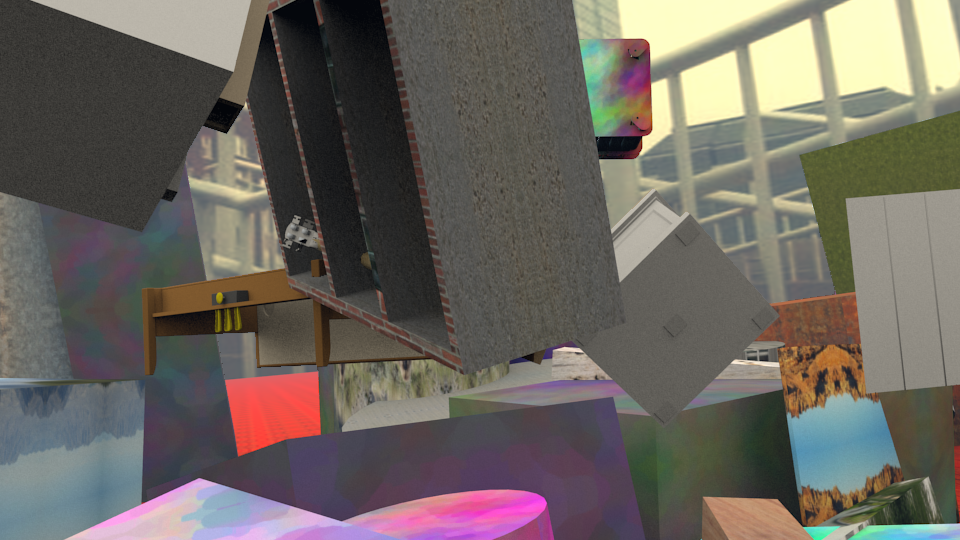}
\end{minipage}%
}%
\subfigure[Stacked GRU~\cite{yao2019recurrent}]{
\begin{minipage}[b]{0.24\textwidth}
\centering
\includegraphics[width=\textwidth]{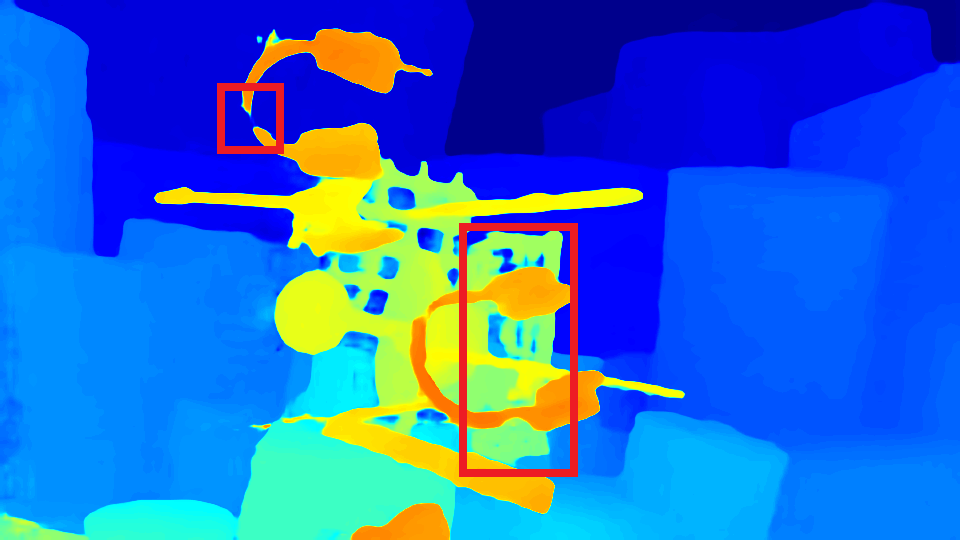}
\includegraphics[width=\textwidth]{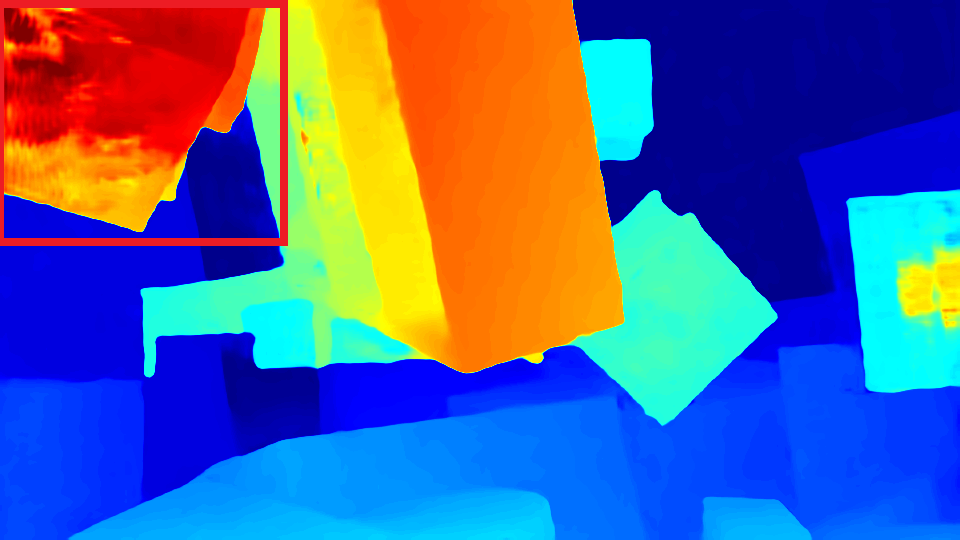}
\end{minipage}%
}%
\subfigure[SRH-Net]{
\begin{minipage}[b]{0.24\textwidth}
\centering
\includegraphics[width=\textwidth]{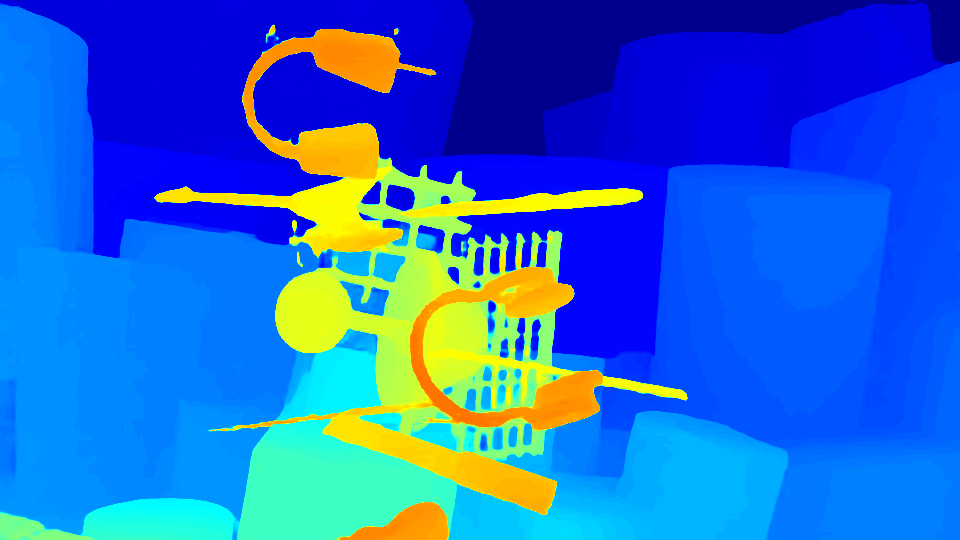}
\includegraphics[width=\textwidth]{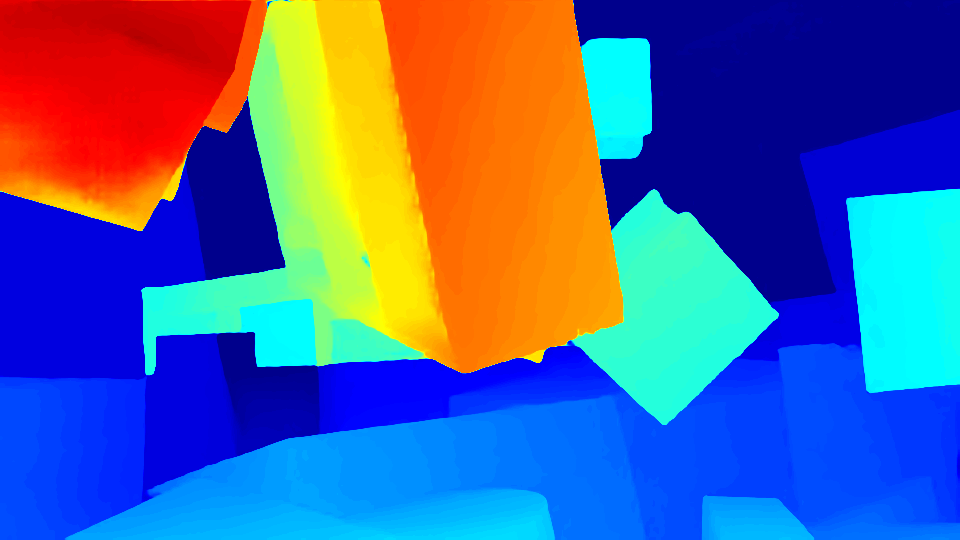}
\end{minipage}
}%
\subfigure[Ground Truth]{
\begin{minipage}[b]{0.24\textwidth}
\centering
\includegraphics[width=\textwidth]{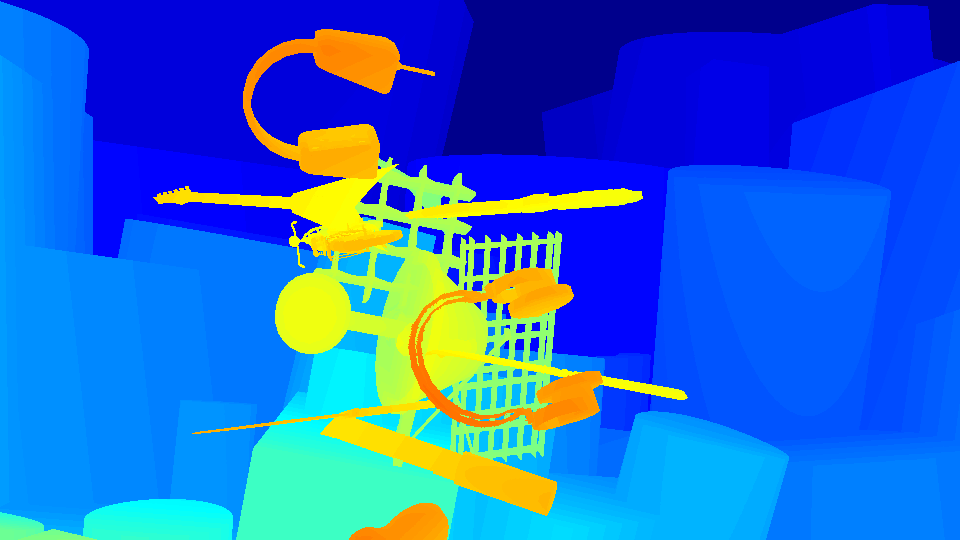}
\includegraphics[width=\textwidth]{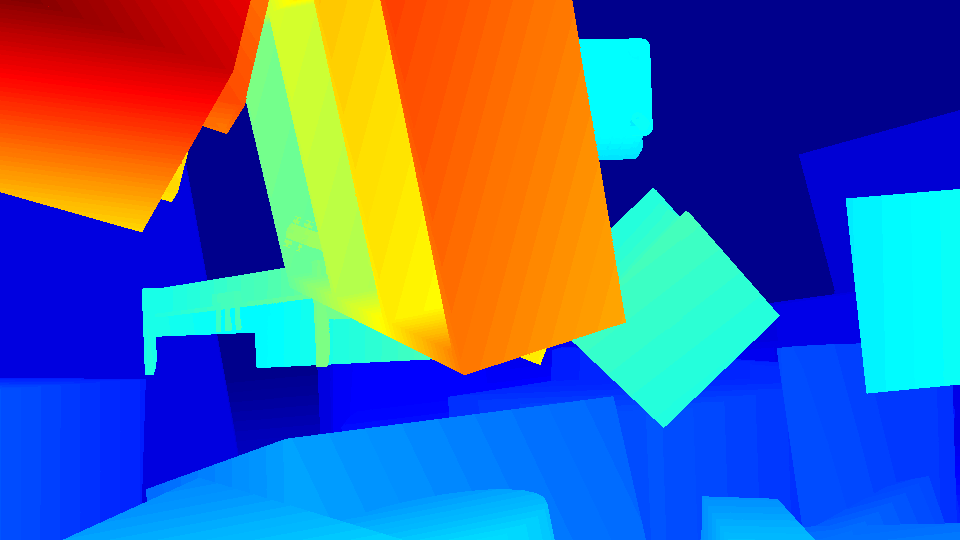}
\end{minipage}
}%
\centering
\caption{Comparison of results from different cost aggregation modules on the Scene Flow datasets. The memory consumption of our architecture is only 2.42 GB which is reduced by up to 56.1\% compared with PSMNet~\cite{chang2018pyramid} using stacked hourglass 3D CNNs without the degradation of accuracy. }
\label{fig:scene_result}
\end{figure*}

\subsection{Training loss}
After collecting aggregated cost maps, a disparity volume is constructed by upsampling those maps to $\text{H}\times \text{W}\times \text{D}$ via bilinear interpolation. Then the entire volume is regressed to a continuous 2D disparity map $\hat{d}$ using the soft argmin operation~\cite{kendall2017end},
\begin{equation}
\label{equ:regression}
\hat{d} = \sum_{i=0}^{D}i \times softmax(-C_{r}(i))
\end{equation}
where $softmax(\cdot)$ is the softmax operation. In this way, the cost volume is converted to a probability map normalized along the disparity direction. The disparity is finally computed from the expectation value to replace the argmax operation (not differentiable), and produce sub-pixel disparity estimates. After the regression, we adopt the smooth $L_1$ loss function~\cite{R2015fastrcnn} to supervise the predicted disparity map with the ground truth disparity map $d_{gt}$, 
\begin{equation}
L\left( \hat{d},d_{gt} \right)=\frac{1}{N}\sum\limits_{n=1}^{N}{l\left( \left| \hat{d}-d_{gt} \right| \right)}
\end{equation}
here $
l\left( x \right)=\left\{ \begin{aligned}
  & x-0.5,    x\ge 1 \\ 
 & {{{x}^{2}}}/{2,       x<1}\; \\ 
\end{aligned} \right.
$ and $N$ represents the number of labeled pixels.

As mentioned in Section~\ref{section:recurrent aggregation}, we obtain two aggregated cost maps at the $i$ disparity displacement, i.e. intermediate map ($\hat{d_m}$) and ultimate map ($\hat{d_f}$), during the training process, which are regressed to disparity maps. 
Therefore, the final loss function of SRH-Net is a weighted sum of two losses:
\begin{equation}
\label{equ:finalloss}
Loss={{w}_{1}}{L}(\hat{d_m},d_{gt})+{{w}_{2}}L(\hat{d_f},d_{gt})
\end{equation}
where $w_1$ and $w_2$ are the weights for the two-stage output.

\section{EXPERIMENTS}\label{experiments}
To evaluate the proposed SRH-Net, we test it on popular public benchmarks such as Scene Flow~\cite{mayer2016large}, KITTI (2012~\cite{genger2012KITTI}, 2015~\cite{Menze2015ISA}) and Middlebury2014~\cite{2014High}. In the ablation studies, the SRH module, stacked GRU and stacked 3DCNN hourglass are merged into the same architecture respectively, which provides the same cost map construction and disparity regularity modules. Then, we compare our entire pipeline with other state-of-the-art models in terms of GPU memory consumption and disparity prediction accuracy.

\subsection{Experimental settings}
\paragraph{\textbf{Datasets}}
A stereo matching model needs sufficient image pairs with reliable depth for training. However, it is difficult to collect accurate and dense disparity maps in many practical scenarios. Therefore, the popular way is to train models on synthetic data first and fine-tune them on a small number of real pairs, which improves the performance of disparity estimation in real scenes and avoid overfitting issues simultaneously.

Scene Flow is a large dataset with 39,824 samples of synthetic stereo RGB images in $960\times540$ pixels resolution. For each synthetic sequence, it provides high-quality dense disparity maps as ground truth while there are 4370 image pairs with different rendered assets for testing. In our experiments, end-point-error (EPE) and 1-pixel error rate (1-ER) are computed in the range of our disparity estimation for evaluation.
Different from Scene Flow, the KITTI dataset is a real-world dataset collected in road scenes, which consists of KITTI 2012~\cite{genger2012KITTI} and KITTI 2015~\cite{Menze2018JPRS,Menze2015ISA}. These subsets provide 194 and 200 training stereo pairs in $1249\times376$ pixels resolution separately. The ground truth includes sparse disparity maps generated by LiDAR. The results of our visualization and evaluation on the KITTI dataset are obtained from its online leaderboard. 

\paragraph{\textbf{Implementation}}
We implement our SRH-Net on the PyTorch framework, optimized via Adam optimizer ($\beta_{1}=0.9, \beta_{2}=0.999$). We normalize inputs by subtracting the means and dividing standard deviations of each channel. Due to the limitations of our GPUs (a RTX 2080Ti and a RTX Titan), the input images are randomly cropped to a size of 240$\times$576 pixels, and the batch size is set to 3. We set the maximum disparity (D) to 192 and
the size of feature maps is $1/4H\times1/4W\times32$ where $H$ and $W$ represent the height and width of the input images.
For the Scene Flow dataset, the model is directly used for the evaluation after 10-epoch training with a constant learning rate of 0.001. For fine-tuning on KITTI subsets, we use the pretrained model trained on the Scene Flow dataset, which is fine-tuned through 800 epochs in each subset. The learning rate is set to 0.001 for the first 400 epochs and decreased to 0.0001 for the remaining ones.

\subsection{Ablation study for loss weight}
As shown in Table~\ref{tab:weights}, we compare with various combinations of loss weights for the intermediate supervision after 10-epoch training. 
First, the intermediate supervision is removed by setting $w_1=0$ and different rates of two parameters are exploited by changing $w_2$. When the weights ($w_1$, $w_2$) in the loss function~\eqref{equ:finalloss} are set to $0.4$ and $1.2$, the system yields the best performance on the Scene Flow test set.

\begin{table}[h]
\centering
\caption{Comparison of different weights for intermediate supervision on the Scene Flow test set for 10 epochs. EPE means average end-point-error and 1-ER is the percentage of outliers greater than 1 pixel.}
\setlength{\tabcolsep}{0.05\textwidth}{
\begin{tabular}{c|c|c|c|c}
\hline
\multicolumn{3}{c|}{Loss Weights} & \multirow{2}{*}{EPE}  & \multirow{2}{*}{1-ER(\%)}\\ \cline{1-3}
$w_1$               & $w_2$  & rate            &                 &                      \\ \hline
0.4              & 0.4      & 1:1       & 1.31                  & 13.2                    \\ \hline
0.4              & 0.8      & 1:2       & 1.21                  & 11.8                    \\ \hline
0.4              & 1.2      & 1:3       & \textbf{1.09}                 & \textbf{10.1}  \\ \hline
0.0              & 1.0      & 0:1       & 1.25                  & 12.4                  \\ \hline
\end{tabular}
}
\label{tab:weights}
\end{table}
\subsection{Ablation study for cost aggregation}
\label{ablation study}
In order to evaluate the performance of our SRH module, we compare it with other popular cost aggregation strategies within the same pipeline. Figure~\ref{fig:example} and Table~\ref{table:ablation_aggregation} report the results of the comparison between the following modules:
\begin{itemize}
    \item \textbf{Stacked 3DCNN hourglass} is proposed in PSM-Net~\cite{chang2018pyramid}, which starts from two 3D convolutional layers with a residual block. And then, three hourglass networks and two 3D convolutions are exploited to reduce the number of channels to 1. All 3D convolutional layers in this module are implemented with the size of 3$\times$3$\times$3. 
    \item \textbf{Stacked GRU} is proposed in RMVSNet~\cite{yao2019recurrent}.The stacked GRU module used here has a 3-layers stacked 3$\times$3 convolutional GRUs with 32-channel outputs, followed by two 3$\times$3 convolutional layers to generate 1-channel cost maps.
\end{itemize}
The details of SRH-Net architecture are described in section~\ref{section:recurrent aggregation}. From the feature extraction and warping module, a 4D cost volume (or sequence) with the size of $1/4H\times 1/4W\times 1/4D\times 64$ is created for each cost aggregation method. 

\begin{table}[h]
\centering
\caption{Comparison of results with different cost aggregation modules.}
\begin{spacing}{1.2}
\begin{tabular}{lccc}
\hline
Methods  & EPE(pixel) & 1-ER(\%)  & Memory(GB) \\ \hline
3DCNN                      & \textbf{1.09}            & 12.1                    & 4.98  \\
Stacked GRU                 & 1.66            & 16.3                    & \textbf{1.99}    \\
\textbf{SRH (Ours)} & \textbf{1.09}            & \textbf{10.1}                    & 2.42   \\ \hline
\end{tabular}
\end{spacing}
\label{table:ablation_aggregation}
\end{table}

As listed in Table~\ref{table:ablation_aggregation}, different cost aggregation modules are evaluated on the Scene Flow dataset. 3DCNN and our module obtain the best average EPE of 1.09 pixel, though the proposed module achieves the best 1-pixel threshold error rate of 10.1\%, i.e. better than the other two methods. The fourth column shows that GRU-based methods can reduce GPU memory consumption significantly. Compared with Stacked GRU, 3DCNN takes 1.8 times of GPU memory for estimating an equally sized disparity map, but yields the same performance with SRH-Net at 1.09 pixels in terms of average EPE. Furthermore, compared with stacked GRU, 22\% extra GPU memory is required by SRH-Net, but SRH-Net improves the average EPE by 38\%. These results indicate that our SRH module improves performance by using the stacked hourglass structure. Hence the module can represent an efficient solution for deployment on generic mobile devices.

In order to observe and analyze the improvement from the SRH module, disparity maps are compared in Figure~\ref{fig:example}. As illustrated in the red boxes, our SRH-Net performs more robust disparity estimation in textureless and occluded regions than the stacked GRU method since it fails to implement a larger receptive field. For such challenging regions, it is difficult to determine the correspondences only from a fixed scale. Therefore, these results can be used to illustrate that the proposed SRH module is good at capturing contextual information from each level of the cost pyramid.

\subsection{Comparison with the state-of-the-art methods}
\begin{table}[t]
\vspace{6pt}
\small
\centering
\caption{Comparison with other methods. \textit{w} represents that the method contains 3D convolutions, while \textit{wo} represents that the methods only use 2D convolutions. `Noc' and `All' represent non-occluded and all regions respectively. `D1-bg' and `D1-all' indicate the percentage of outliers averaged over background and all ground truth pixels respectively. }
\begin{tabular}{cc|cc|cc}\hline
\multicolumn{2}{c|}{\multirow{2}{*}{Methods}} & \multicolumn{2}{c|}{All} & \multicolumn{2}{c}{Noc} \\ 
\multicolumn{2}{c|}{}                   & D1-bg   & D1-All & D1-bg  & D1-All \\ \hline \hline
DispNetC~\cite{mayer2016large}                 & \textit{wo} & 4.32      & 4.34  & 4.11     & 4.05          \\
CRL~\cite{pang2017cascade}                      & \textit{wo} & 2.48     & 2.67   & 2.32   & 2.45         \\
FADNet~\cite{wang2020fadnet}                   & \textit{wo} & 2.68     & 2.82  & 2.49   & 2.59           \\
AANet~\cite{AANet}                    & \textit{wo} & 1.99     &2.55   &1.80    &2.32\\
\textbf{Ours}            & \textit{wo} & \textbf{1.86}   & \textbf{2.26} & \textbf{1.70} &\textbf{2.05} \\ \hline
GC-Net~\cite{kendall2017end}                    & \textit{w}  & 2.21    & 2.87 & 2.02    & 2.61          \\
PSMNet~\cite{chang2018pyramid}                   & \textit{w}  & 1.86   & 2.32   & 1.71    & 2.14           \\
GA-Net~~\cite{Zhang2019GANet}                    & \textit{w} & \textbf{1.48}  & \textbf{1.81} & \textbf{1.34}  & \textbf{1.63} \\ \hline 
\end{tabular}
\label{table:comparision}
\end{table}

\begin{table}[t]
\begin{tabular}{c|ccc}
\hline
Methods & Parames & Memeory &Runtime \\ \hline
PSMNet  &5.2 M    & 4.65 G    &0.4 s       \\
GANet   &6.5 M   & 6.35 G    & 2.2 s      \\
Ours    &4.7 M   & 2.04 G   &0.5 s      \\ \hline
\end{tabular}
\caption{Comprehensive comparison between networks on the KITTI datasets.}
\label{table:compare_memory}
\end{table}

\setlength{\tabcolsep}{0.021\textwidth}{
\begin{table*}[t]
\vspace{6pt}
\centering
\caption{Results on the KITTI2012 dataset. The metrics present the percentage of bad pixels (estimated error from $>2$ to $>5$ pixels) and 'Avg. error' is the average disparity error (pixel).}
\begin{tabular}{c|c|cccccccccc}
\hline
\multicolumn{2}{c|}{\multirow{2}{*}{Methods}} & \multicolumn{2}{l}{$>2$ px}          & \multicolumn{2}{l}{$>3$ px}          & \multicolumn{2}{l}{$>$4 px}          & \multicolumn{2}{l}{$>5$ px}          & \multicolumn{2}{l}{Avg. error}        \\ \cline{3-12} 
\multicolumn{2}{c|}{}               & Noc           & All           & Noc           & All           & Noc           & All           & Noc           & All           & Noc          & All          \\ \hline
DispNetC~\cite{mayer2016large}&\textit{wo}                    & 7.38          & 8.11          & 4.11          & 4.65          & 2.77          & 3.20          & 2.05          & 2.39          & 0.9          & 1.0          \\
FADNet~\cite{wang2020fadnet}  &\textit{wo}              & 3.98          & 4.63          & 2.42          & 2.86          & 1.73          & 2.06          & 1.34          & 1.62          & 0.6          & 0.7          \\
AANet~\cite{AANet}   &\textit{wo}             &2.90            &3.60           &1.91           &2.42          &1.46            &1.87           &1.20           &1.53          &0.5         &0.6\\
\textbf{Ours}    &\textit{wo}            & \textbf{2.07}          & \textbf{2.64}          & \textbf{1.27}          & \textbf{1.66}          & \textbf{0.94}          & \textbf{1.25}          & \textbf{0.75} & \textbf{1.00} & \textbf{0.5}          & \textbf{0.5} \\ \hline
GC-Net~\cite{kendall2017end}  &\textit{w}                 & 2.71          & 3.46          & 1.77          & 2.30          & 1.36          & 1.77          & 1.12          & 1.46          & 0.6          & 0.7          \\
PSMNet~\cite{chang2018pyramid}   &\textit{w}               & 2.44          & 3.01          & 1.49          & 1.89          & 1.12          & 1.42          & 0.90          & 1.15          & 0.5          & 0.6          \\
GA-Net~\cite{Zhang2019GANet}    &\textit{w}                & \textbf{1.89} & \textbf{2.50} & \textbf{1.19} & \textbf{1.60} & \textbf{0.91} & \textbf{1.23} & 0.76          & 1.02          & \textbf{0.4} & \textbf{0.5} \\
 \hline
\end{tabular}
\label{table:compare_k12}
\end{table*}}

We compare our SRH-Net with other models on the KITTI datasets in Table~\ref{table:comparision} and \ref{table:compare_k12}, where 'D1' refers to the percentage of pixels with errors more than 3 pixels or 5 of disparity error from all test images and 'Avg. error' refers to the average of end-point-errors. All metrics are evaluated in non-occluded (Noc) and all (All) regions, respectively. Here the 2DCoder methods include DispNetC~\cite{mayer2016large}, FADNet~\cite{wang2020fadnet} and CRL~\cite{pang2017cascade}, while GC-Net\cite{kendall2017end}, PSMNet~\cite{chang2018pyramid} and GA-Net~\cite{Zhang2019GANet} belong to 3DConv approaches. To present the results more intuitively, DispNetC and GC-Net are selected as the baselines of those two classes, respectively. Furthermore, similar to our method, AANet~\cite{AANet} makes use of 2D operations in the 3DConv framework.
 
\textbf{Comparison with 2DCoder methods.} As it can be seen in Table~\ref{table:comparision}, 2DCoder methods generally perform worse than GC-Net in non-occluded regions, which shows that an explicit cost aggregation step is useful to improve networks' performance in stereo matching. Although CRL and FADNet have a disparity refinement process, they are still slightly worse than PSMNet over all regions. Our network obtains more robust results on those two datasets, improving of more than 35\% compared to DispNetC. However, CRL and FADNet are too sensitive to non-occluded regions of KITTI 2015, which degenerate to 43\% and 25\% compared to our method, respectively. Furthermore, on KITTI 2012, we observe a 48\% improvement of the 3-pixel error rate compared with FADNet for non-occluded areas, and our method also performs better in other metrics.

\textbf{Comparison with 3DConv methods.} As shown in Table~\ref{table:comparision} and~\ref{table:compare_k12}, GA-Net~\cite{Zhang2019GANet} achieves state-of-the-art results on the KITTI datasets by using the semi-global and local aggregation layers to guide 4D cost aggregation. Moreover, we also compare the number of parameters and running time of those methods as shown in Table~\ref{table:compare_memory}. However, numerous computational resource is also required by GA-Net where GPU memory consumption increases by 3.1 times compared to our approach, i.e. from 2.04 GB to 6.35 GB. PSMNet and our network yield the same disparity estimation accuracy at 1.09 average EPE on the scene flow dataset, though our SRH-Net obtains better performance in KITTI 2015 and KITTI 2012. Furthermore, our method requires only 2.04 GB, i.e. a drop in memory consumption of 56.1\% compared to PSMNet.

\begin{figure}[t]
\vspace{6pt}
  \centering
   \includegraphics[width=0.9\textwidth]{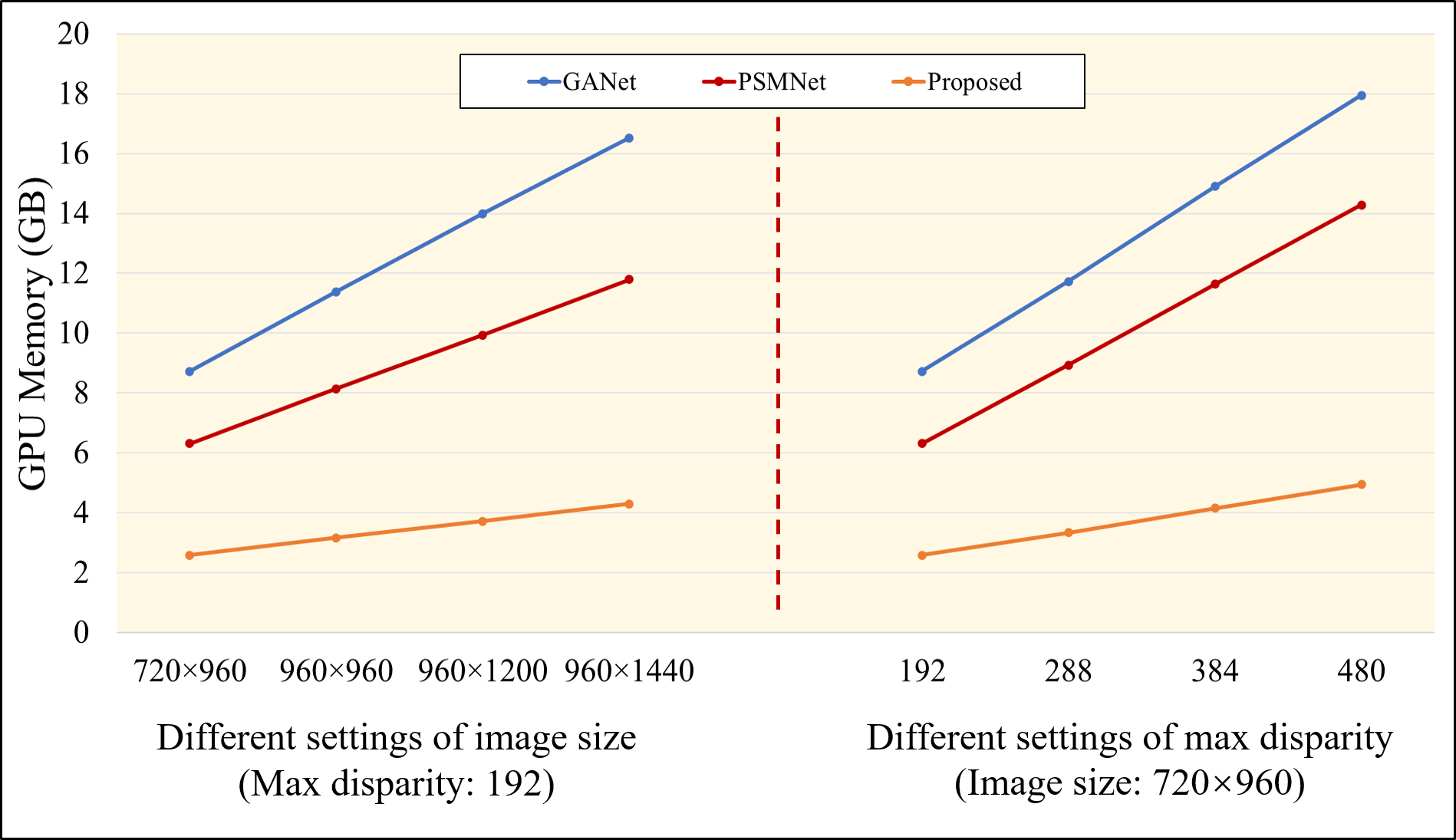}
   \caption{Comparison of memory consumption with different settings: the resolution (left) and max disparity (right).}
      \label{fig:chat}
\end{figure}

For 3DCoder methods, the requirements of computation resources are determined by the size of images and the range of disparity estimation, which are regarded as metrics to evaluate the consumption of GPU memory between those state-of-the-art methods. The comparison is implemented on a Nvidia Titan RTX GPU (24GB) with the original settings proposed by each paper. As shown in Figure~\ref{fig:chat}, SRH-Net has a few increments of memory consumption during the increasing of the image sizes and maximum disparity, which suggests that SRH-Net can suffice for the disparity estimation of large-size images without down-sampling or tiling.

\begin{figure*}[t]
  \centering
  \subfigure[Inputs]{
  \includegraphics[width=0.23\textwidth]{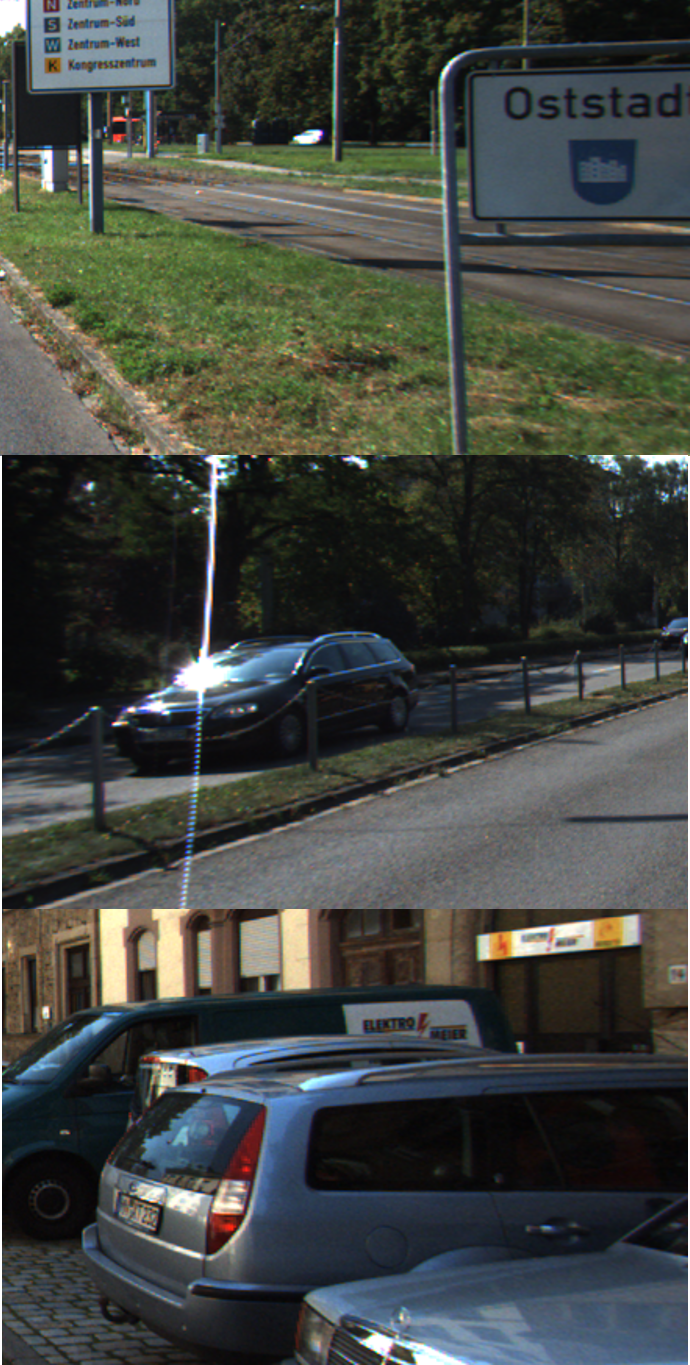}
}
    \subfigure[AANet~\cite{AANet}]{
    \includegraphics[width=0.23\textwidth]{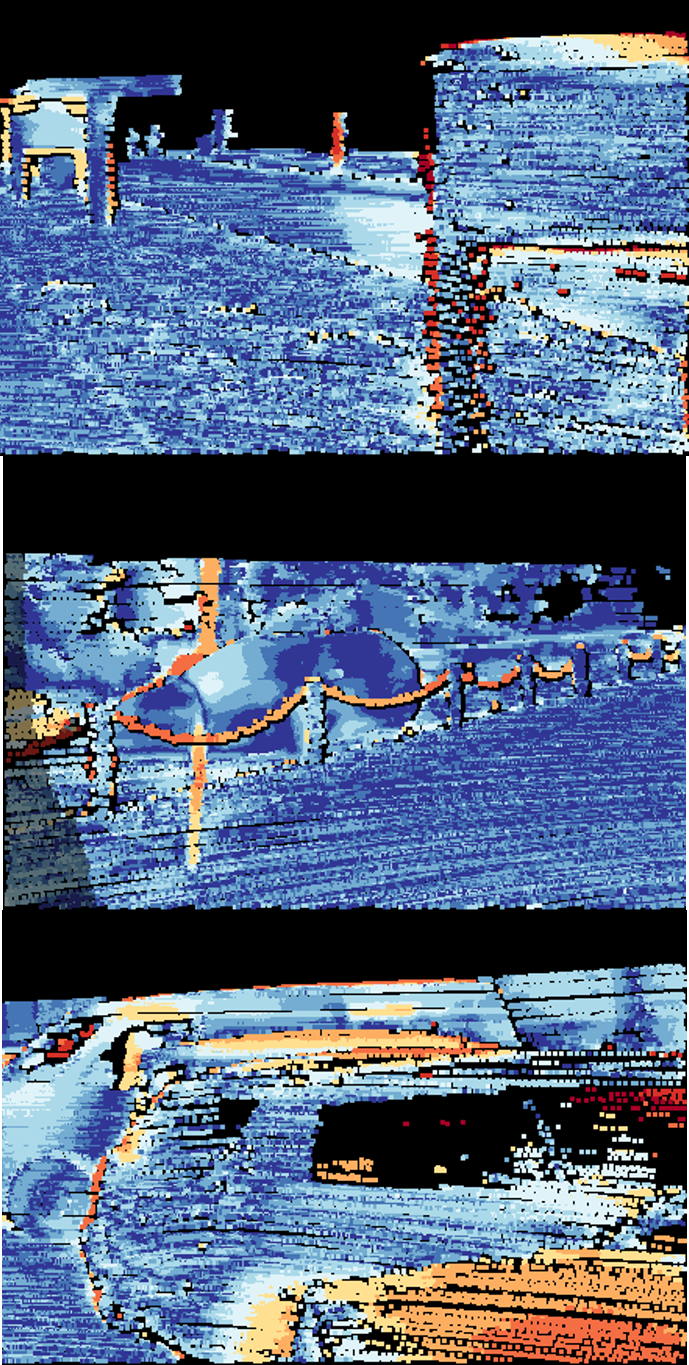}}
     \subfigure[PSMNet~\cite{chang2018pyramid}]{
    \includegraphics[width=0.23\textwidth]{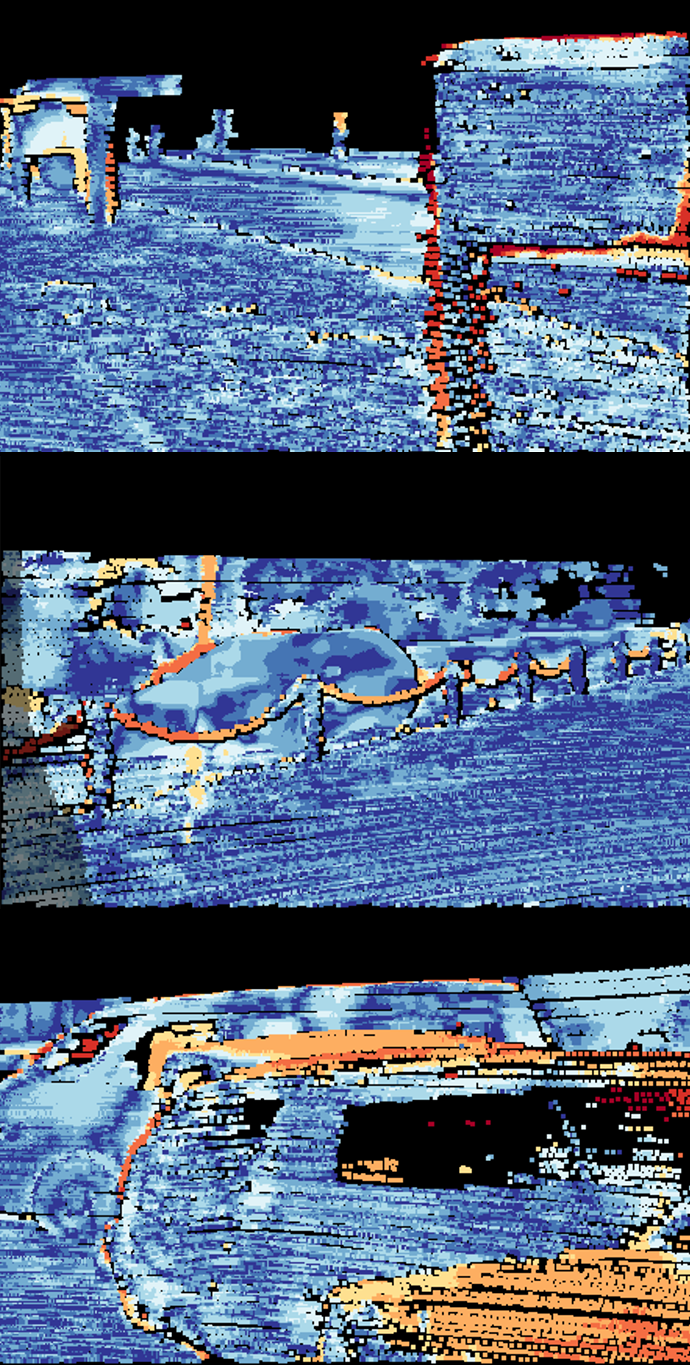}} 
    \subfigure[SRH-Net]{
    \includegraphics[width=0.23\textwidth]{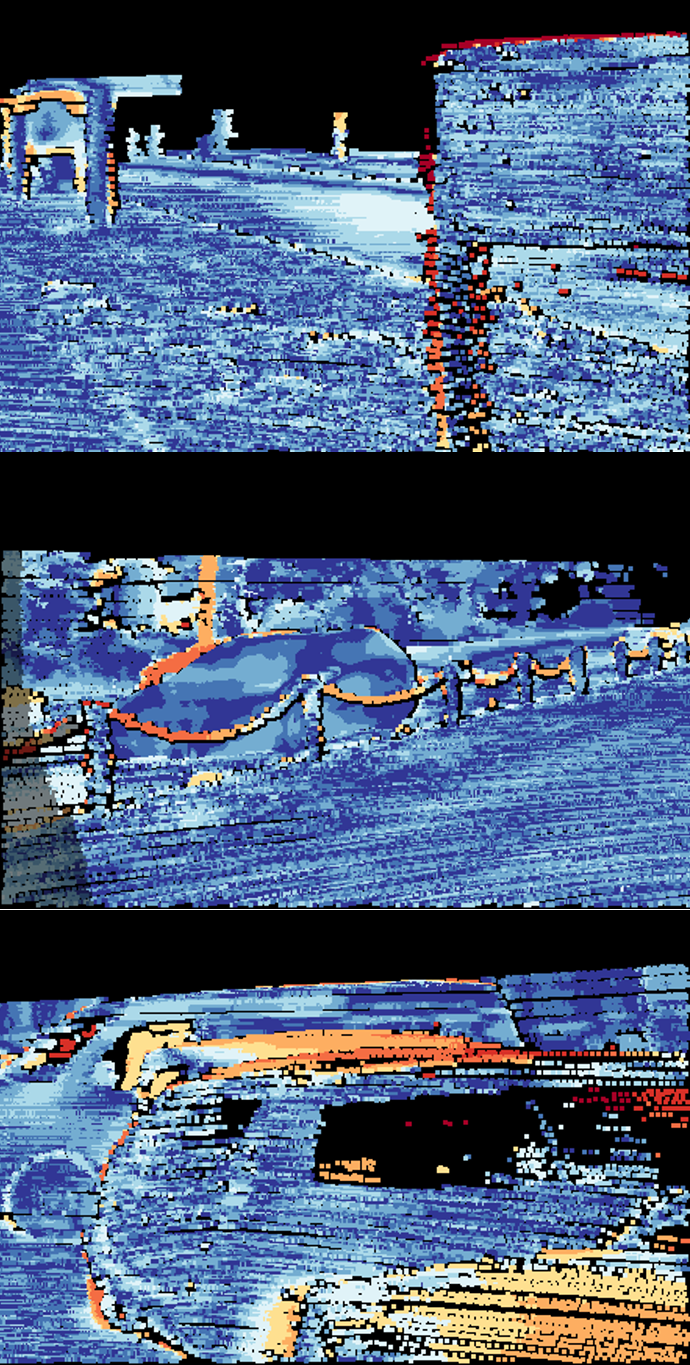} \ } 
   \caption{Comparison with the KITTI 2015 testing images. For each method, the error maps are illustrated where the bad estimates are colored in red.}
      \label{fig:kitti15}
\end{figure*}

\textbf{Qualitative evaluation.} As shown in Figure~\ref{fig:kitti15}, we compare our results with AANet and PSMNet in terms of error maps. 
As visible from the first row, the proposed SRH-Net performs better compared with AANet and PSMNet, especially in the low-textured objects. In the last two rows, reflective areas can be found in the input images, where SRH-Net shows robust performance in those areas.
Although AANet can be enhanced by taking a powerful refinement module~\cite{AANet,2019StereoDRNet}, the pipeline still cannot avoid domain shift between different datasets. These results suggest that the low-dimensional cost representation in AANet breaks the integrity of information, which performs less robustly than other methods using concatenation volume.

\begin{figure}[t]
\vspace{6pt}
  \centering
  \subfigure[Inputs]{
  \includegraphics[width=\textwidth]{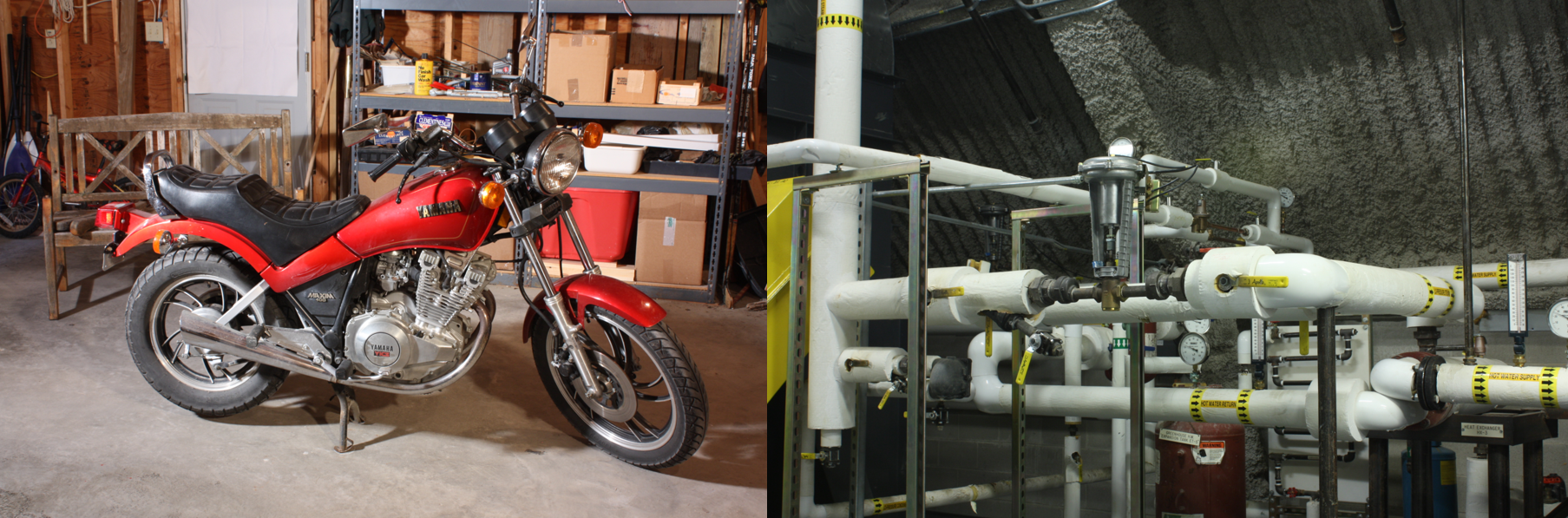}
}\vspace{-5pt}
    \subfigure[AANet~\cite{AANet}]{
    \includegraphics[width=\textwidth]{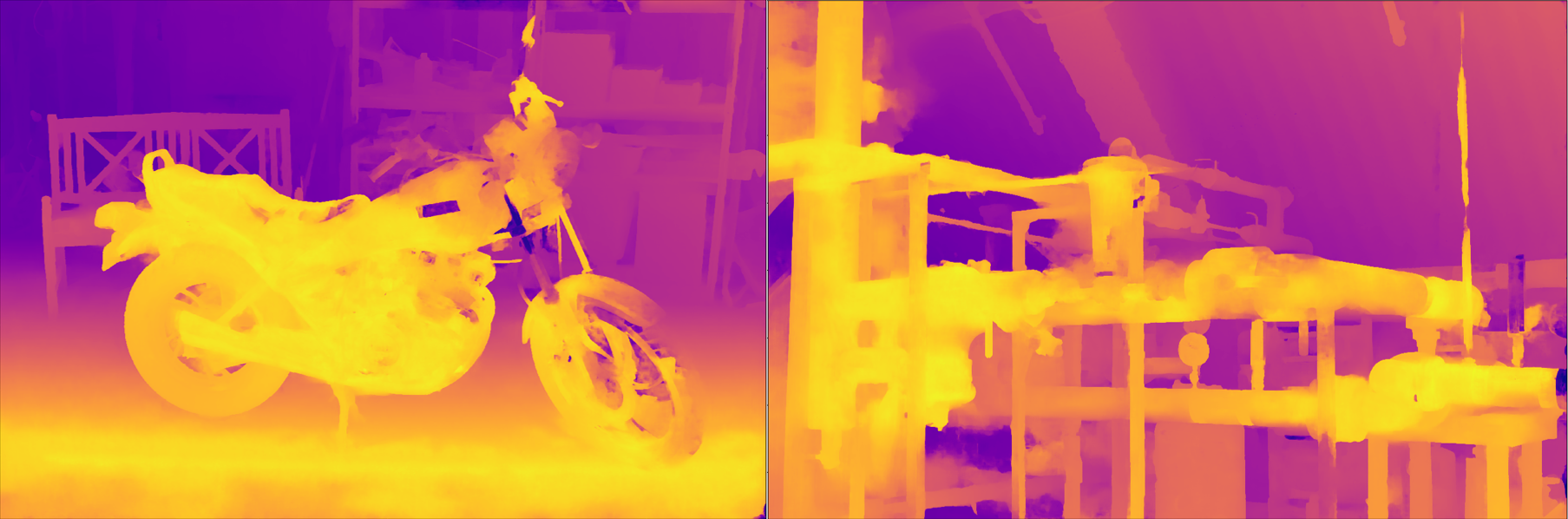}}\vspace{-5pt}
    \subfigure[SRH-Net]{
    \includegraphics[width=\textwidth]{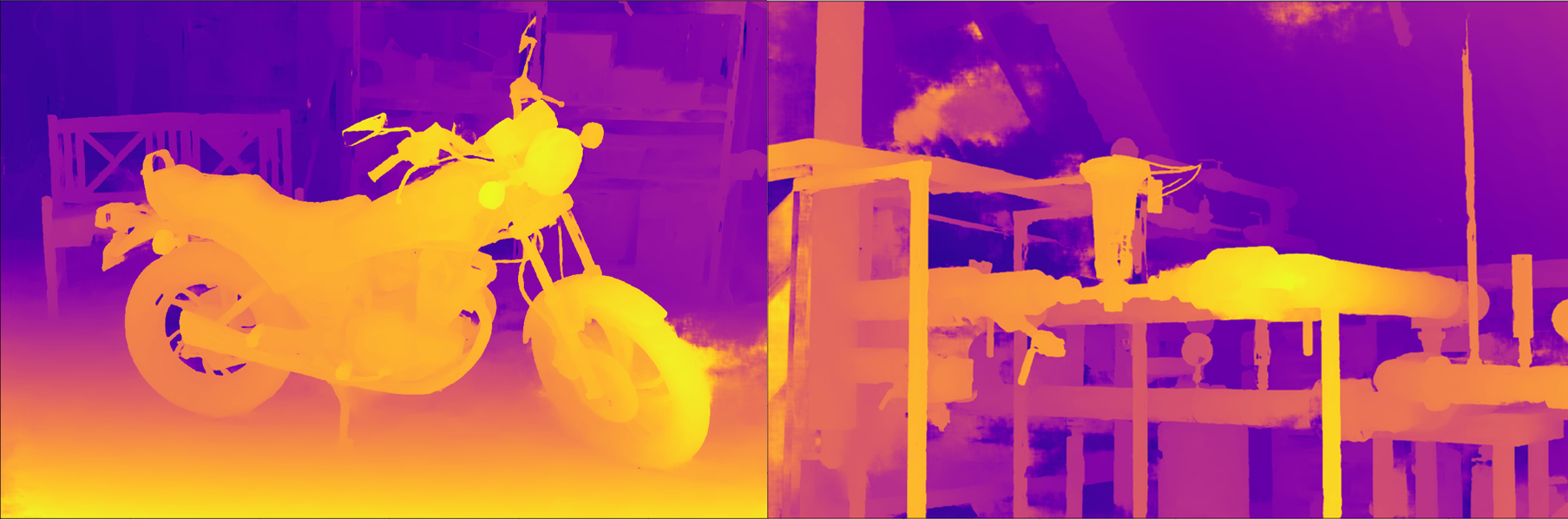}\ }
   \caption{Samples for high-resolution disparity results. The image sizes are 2960$\times$1920 pixels (left) and 2960$\times$2000 pixels (right), respectively.}
      \label{fig:generalization}
\end{figure}
\subsection{Validation of generalization}
To evaluate the generalization of the model trained on scene flow, Middlebury2014~\cite{2014High} is used to test different methods, which provides high-resolution stereo image pairs recorded from different scenarios.
As shown in Figure~\ref{fig:generalization}(a), the resolution of those two pairs are $960\times 2000$ and $2960\times 1924$, respectively. When we feed those stereo pairs to PSMNet~\cite{chang2018pyramid} and GANet~\cite{Zhang2019GANet}, those models cannot run on a machine with limited memory, like 24G, directly. Different from 3DConv methods, AANet needs less intensive computation since it makes use of 2D convolutions to aggregate the 3D cost volume, which obtains a 0.87 EPE by using only 1.4G memory on the Sceneflow dataset. The max disparity of the model, however, is fixed in the architecture, leading to bad performances in challenging regions.
Compared with AANet, our model shares the weights for each disparity level during the cost aggregation, which can be used to take place of 3DConv methods to estimate high-resolution disparity maps directly, rather than extra training with different settings. More qualitative results are shown in the video.

\section{CONCLUSIONS}

In this paper, an efficient recurrent cost aggregation strategy is proposed to estimate accurate disparity maps within limited GPU consumption. This strategy is implemented in an end-to-end pipeline based on a novel module dubbed SRH. 
In the cost aggregation stage, the SRH module generates multi-scale cost maps at each disparity level, which makes the network more robust under challenging scenes by capturing more structural and contextual information.
Compared with stacked GRUs, our SRH module provides more robust costs for disparity computation, especially in textureless and occluded regions. Furthermore, the performance of SRH-Net is evaluated on public datasets, where the proposed method yields competitive predictions compared with state-of-the-art architectures and satisfies the application needs of high-resolution reconstruction.

\bibliography{IEEEexample}

\end{document}